# PSLF: A PID Controller-incorporated Second-order Latent Factor Analysis Model for Recommender System

Jialiang Wang
*College of Computer and Information Science*
*Southwest University*
Chongqing, China
goallow@163.com

Yan Xia
*School of Electronic Information and Communication Engineering*
*Chongqing Aerospace Polytechnic*
Chongqing, China
xiayan_99@163.com

Ye Yuan
*College of Computer and Information Science*
*Southwest University*
Chongqing, China
yuanyekl@swu.edu.cn

*Abstract*—A second-order-based latent factor (SLF) analysis model demonstrates superior performance in graph representation learning, particularly for high-dimensional and incomplete (HDI) interaction data, by incorporating the curvature information of the loss landscape. However, its objective function is commonly bi-linear and non-convex, causing the SLF model to suffer from a low convergence rate. To address this issue, this paper proposes a PID controller-incorporated SLF (PSLF) model, leveraging two key strategies: a) refining learning error estimation by incorporating the PID controller principles, and b) acquiring second-order information insights through Hessian-vector products. Experimental results on multiple HDI datasets indicate that the proposed PSLF model outperforms four state-of-the-art latent factor models based on advanced optimizers regarding convergence rates and generalization performance.

I. INTRODUCTION

In a recommender system, each user and item interaction can be labeled as a rating, describing the user's preference for an item. However, with the rapid development of users and items scale, it is difficult for each user to rate all items, and likewise, it is difficult for each item to be rated by all users. High-dimensional and incomplete (HDI) rating matrices are widely adopted to represent this high-dimension and sparse interaction behavior in recommender system [1-22, 108, 109].

Each row of the HDI matrix represents a user, each column represents each item, and the matrix's values, the ratings $r_{u,i}$, describe the preference for user $u$ to item $i$. Albeit the interaction behaviors are sparse, i.e., the HDI matrix includes many null elements, we can extract latent knowledge from these sparse interaction behaviors to obtain their data representation.

The latent factor analysis (LFA) is a graph representation learning model used for representing HDI data. The LFA model assumes that each user's and each item's features can be mapped into a low-rank latent factor matrix (aka embedding space). The rating related to user $u$ and item $i$ can be obtained via the inner product between $u$'s latent factor vector and $i$'s latent factor vector. Hence, the learning objective of the LFA model is bi-linear and non-convex.

To achieve optimal representation ability of the LFA model while balancing multiple aspects such as computation complexity, storage complexity, and missing data prediction performance. There are three types of optimization-based LFA models [23], i.e., first-order-based LFA [24-28], adaptive gradient-based LFA [29-32], and second-order-based LFA [33-39, 73]. Among them, the second-order-based LFA outperforms the first-order-based LFA and adaptive gradient-based LFA in representation ability since it incorporates curvature information of the LFA model's objective function at a given point.

Commonly the second-order optimizer solve the linear system as follows:

$$g_E(\mathbf{X}) + \mathbf{H}_E(\mathbf{X})\Delta\mathbf{X} = 0, \qquad (1)$$

where $\mathbf{X} \in \mathbb{R}^d$ is the decision parameter vector, $\Delta\mathbf{X} \in \mathbb{R}^d$ denotes the incremental vector, $g_E(\mathbf{X}) \in \mathbb{R}^d$ and $\mathbf{H}_E(\mathbf{X}) \in \mathbb{R}^{d \times d}$ denotes the gradient vector and Hessian matrix w.r.t objective function $E(\mathbf{X})$, respectively. However, the second-order optimizers, e.g., Newton-type methods, require $O(d^2)$ to compute and store $\mathbf{H}_E(\mathbf{X})$, and $O(d^3)$ to compute $\mathbf{H}_E(\mathbf{X})$'s inverse [33-39, 73]. Although the second-order optimizers theoretically converge faster compared to first-order optimizers, the expensive computational and storage costs hinder the practicality of second-order optimizers in handling large-scale data. It has been found that we can obtain the incremental vector $\Delta\mathbf{X}$ by computing the Hessian-vector product across multiple conjugate gradient iterations. This means that we do not need to manipulate the Hessian matrix $\mathbf{H}_E(\mathbf{X})$ and its inverse directly, thus mitigating the curse of dimensionality. However, the Hessian-vector-based LFA model requires multiple iterations to converge in some data scenarios [33-39, 73].

As in previous research [41-58], incorporating the principle of PID controller to refine stochastic gradient and learning error, which can accelerate the training process due to acquiring gradient estimation with more accuracy without sacrificing generalization ability. Their ideas are described as:

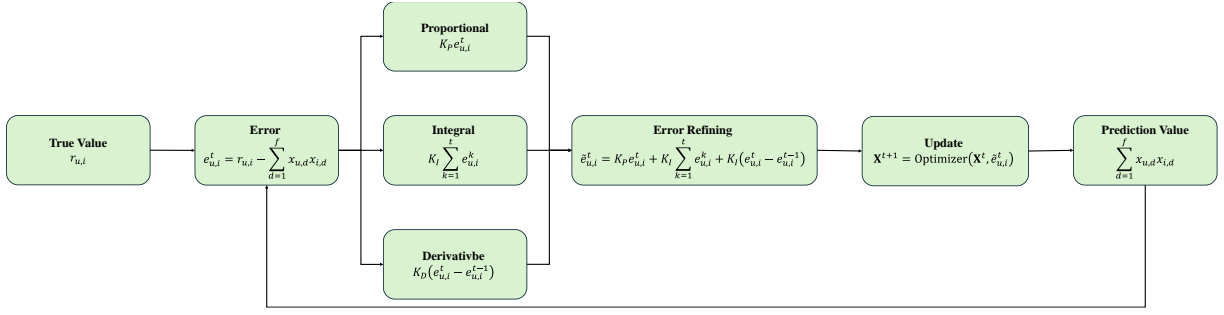

Fig. 1. The scheme of refining learning error via the PID controller.

$$\tilde{e}^t = K_P e^t + K_I \sum_{k=1}^{t} e^k + K_D \left( e^t - e^{t-1} \right), \quad (2)$$

where $e^t$ denotes the true learning error obtained from observed data samples at the $t$-th epoch, $\tilde{e}^t$ denotes the refined/estimated learning error via the PID controller, $K_P$, $K_I$, and $K_D$ are the hyperparameters of the PID controller.

Inspired by the idea of learning error estimated via a PID controller, a <u>P</u>ID controller-incorporated <u>SLF</u> (PSLF) model has been proposed in this paper to obtain a more accurate estimation of gradient vector to accelerate the training process of the SLF model.

Experiment results on four open-access benchmark datasets show that the proposed PSLF model outperforms four state-of-the-art LFA models based on advanced optimizers regarding convergence rate and generalization performance.

The rest of this paper is organized as follows. Section II introduces the preliminary knowledge. Section III presents our proposed method. In Section IV, the experimental results on several datasets are given. Section V provides the conclusion of this paper.

## II. PRELIMINARIES

### A. Problem Statement

The notations of this paper are summarized in Table I.

TABLE I. NOTATIONS

| Parameter | Description |
|---|---|
| $U, I$ | User set and item set. |
| $\mathbf{R}^{|U|\times|I|}, \hat{\mathbf{R}}^{|U|\times|I|}$ | Target HDI matrix and its approximation. |
| $K, M$ | Known and unknown set of $\mathbf{R}$. |
| $\Omega$ | Evaluation set. |
| $\mathbf{X}_U^{|U|\times f}$ | Latent factor matrix of set $U$. |
| $\mathbf{X}_I^{|I|\times f}$ | Latent factor matrix of set $I$. |
| $\mathbf{X}^{(|U|+|I|)\times f}$ | Decision parameter contains $\mathbf{X}_U$ and $\mathbf{X}_I$. |
| $\Delta\mathbf{X}^{(|U|+|I|)\times f}$ | Increment vector of $\mathbf{X}$. |
| $E(\mathbf{X})$ | Objective function of $\mathbf{X}$. |
| $\sigma(\mathbf{X})$ | Implicit function of $\mathbf{X}$. |
| $L(\sigma(\mathbf{X}))$ | Loss function of $\sigma(\mathbf{X})$. |
| $\mathbf{g}_E(\mathbf{X})^{(|U|+|I|)\times f}$ | Gradient vector of $E(\mathbf{X})$. |
| $\tilde{\mathbf{g}}_E(\mathbf{X})^{(|U|+|I|)\times f}$ | Estimated gradient vector of $E(\mathbf{X})$. |
| $\mathbf{H}_E(\mathbf{X})^{((|U|+|I|)\times f)\times((|U|+|I|)\times f)}$ | Hessian matrix of $E(\mathbf{X})$. |
| $\mathbf{H}_L(\mathbf{X})^{((|U|+|I|)\times f)\times((|U|+|I|)\times f)}$ | Hessian matrix of $L(\mathbf{X})$. |
| $\mathbf{G}_E(\mathbf{X})^{((|U|+|I|)\times f)\times((|U|+|I|)\times f)}$ | Gauss-Newton matrix of $E(\mathbf{X})$. |
| $\mathbf{G}_L(\sigma(\mathbf{X}))^{((|U|+|I|)\times f)\times((|U|+|I|)\times f)}$ | Gauss-Newton matrix of $L(\sigma(\mathbf{X}))$. |
| $\mathbf{J}_\sigma(\mathbf{X})^{|K|\times((|U|+|I|)\times f)}$ | Jacobian matrix of $\sigma(\mathbf{X})$. |
| $\mathbf{I}$ | Identity matrix. |
| $\mathbf{1}$ | All-one vector. |
| $\mathbf{v}^{(|U|+|I|)\times f}$ | Conjugate direction in conjugate gradient. |
| $\boldsymbol{\omega}^{(|U|+|I|)\times f}$ | Hessian-vector product vector. |
| $x_{u,d}, x_{i,d}$ | Single latent factor in $\mathbf{X}$. |
| $R\{\cdot\}$ | $R$-operator. |
| $f$ | Latent factor dimension of $\mathbf{X}$. |
| $e^t, e_{u,i}^t$ | Learning error at $t$-th epoch. |
| $\tilde{e}^t, \tilde{e}_{u,i}^t$ | Estimated learning error a $t$-th epoch via PID. |
| $\lambda$ | Regularization coefficient. |
| $\gamma$ | Damping coefficient. |
| $\tau$ | Tolerance coefficient in conjugate gradient. |
| $K_P, K_I, K_D$ | Hyper-parameter of a PID controller. |
| $|\cdot|$ | Computing cardinality of an enclosed set. |

*B. Problem Statement*

The related definitions of this paper are given as follows:

***Definition 1: An HDI Matrix.*** Given two entry sets $U$ and $I$, a target matrix $\mathbf{R} \in \mathbb{R}^{|U| \times |I|}$ organized by multiple tuples, e.g., $(u, i, r)$, each tuple denotes the interaction behavior between $u \in U$ and $i \in I$. Let set $K$ denote the known tuples and set $M$ denote the missing tuples in $\mathbf{R}$, respectively. If and only if the scale of $|K| \ll |M|$, the $\mathbf{R}$ is the so-called HDI matrix.

***Definition 2: An LFA Model.*** Given $U$, $I$ and $K$, a standard LFA model aims to construct a low-rank estimation of $\mathbf{R}$, i.e., $\mathbf{R} \approx \hat{\mathbf{R}} = \mathbf{X}_U \mathbf{X}_I^T$, where $\mathbf{X}_U \in \mathbb{R}^{|U| \times f}$ and $\mathbf{X}_I \in \mathbb{R}^{|I| \times f}$ denote the latent matrix w.r.t $U$ and $I$, respectively, among them symbol $f$ denotes the low-rank dimension of latent space.

To optimize $\hat{\mathbf{R}}$, the objective function of a basic LFA model can be expressed in a form that depends on a single latent factor element, as shown below:

$$E(\mathbf{X}) = \frac{1}{2} \sum_{r_{u,i} \in K} \left( \left( r_{u,i} - \sum_{f=1}^{f} x_{u,d} x_{i,d} \right)^2 + \lambda \sum_{d=1}^{f} \left( x_{u,d}^2 + x_{i,d}^2 \right) \right), \quad (3)$$
$$s.t. \ \forall u \in U, i \in I, d \in \{1,...,f\},$$

where $r_{u,i}$ denotes the $(u, i, r)$-th tuple, $\mathbf{X} \in \mathbb{R}^{(|U|+|I|) \times f}$, a unified decision parameter, contains $\mathbf{X}_U$ and $\mathbf{X}_I$, $x_{u,d}$ denotes the $u$-th row and $d$-th column of $\mathbf{X}_U$, the same definition as $x_{i,d}$, bi-linear term $\sum_{d=1}^{f} x_{u,d} x_{i,d}$ denotes the estimation w.r.t $r_{u,i}$, and $\lambda$ is a hyperparameter controlling the effect of Tikhonov regularization term.

*C. A second-order-based Latent Factor Analysis Model*

An SLF model and other second-order-based graph representation learning models are desired to solve below equation (4) iteratively with a second-order optimization method:

$$\arg \min \mathbf{g}_E(\mathbf{X}) + \mathbf{H}_E(\mathbf{X}) \Delta \mathbf{X}, \quad (4)$$

where $\mathbf{g}_E(\mathbf{X}) \in \mathbb{R}^{(|U|+|I|) \times f}$ denotes the gradient of function $E(\mathbf{X})$, $\mathbf{H}_E(\mathbf{X}) \in \mathbb{R}^{((|U|+|I|) \times f) \times ((|U|+|I|) \times f)}$ represents the Hessian matrix of $E(\mathbf{X})$, $\Delta \mathbf{X} \in \mathbb{R}^{(|U|+|I|) \times f}$ is the increment. Most SLF models adopt the Hessian-free optimization method, computing double Jacobian-vector products in each conjugate gradient iteration, hence, avoiding the manipulation of $\mathbf{H}_E(\mathbf{X})$ and its inverse [33-40, 73].

*D. PID Controller Principles: Refining Learning Error*

Based on prior research [41-58], a representative learning model incorporating the PID controller has been developed to refine errors, adhering to the principles of a PID controller. This approach aims to enhance the understanding of learning errors and gradient estimation, thereby improving convergence rate. The main concept of the PID-incorporated LF model is to capture refinement errors, as summarized by Eq. (5) and depicted in Fig. 1.

$$\begin{cases} e_{u,i}^t = r_{u,i} - \sum_{d=1}^{f} x_{u,d} x_{i,d}, \\ \tilde{e}_{u,i}^t = K_P e_{u,i}^t + K_I \sum_{k=1}^{t} e_{u,i}^k + K_D \left( e_{u,i}^t - e_{u,i}^{t-1} \right), \end{cases} \quad (5)$$

where $\tilde{e}_{u,i}^t$ is the refinement error of $e_{u,i}^t$, constant $t$ represents the $t$-th epoch, $K_P$, $K_I$, and $K_D$ are hyperparameters that govern the impact of the proportional, integral, and derivative terms, respectively.

III. A PID-INCORPORATED SLF MODEL

*A. Reformulation*

To simplify the analysis process of the second-order optimization-based LFA model, this section firstly considers the LF model's loss function of Eq. (1) and maps the bi-linear term $\sum_{d=1}^{f} x_{u,d} x_{i,d}$ into an implicit function $\sigma(\cdot)$. The reformulation form is as follows:

$$\begin{cases} \sigma(\mathbf{X})_{u,i} = \sum_{d=1}^{f} x_{u,d} x_{i,d}, \\ L(\sigma(\mathbf{X})) = \frac{1}{2} \sum_{r_{u,i} \in K} \left( r_{u,i} - \sigma(\mathbf{X})_{u,i} \right)^2, \end{cases} \quad (6)$$

where $\sigma(\mathbf{X})_{u,i}$ is a latent mapping value related to entry $u$ and $i$, $L(\sigma(\mathbf{X}))$ denotes the loss function of Eq. (3).

*B. Gauss-Newton Approximation*

Due to the non-convex nature of loss function $L(\sigma(\mathbf{X}))$, its Hessian matrix $\mathbf{H}_L(\sigma(\mathbf{X}))$ is indefinite. Hence, we adopt the Gauss-Newton matrix $\mathbf{G}_L(\sigma(\mathbf{X}))$, a semi-definite matrix, to approximate its Hessian matrix $\mathbf{H}_L(\sigma(\mathbf{X}))$. Based on prior research [33-39, 73], the Gauss-Newton matrix $\mathbf{G}_L(\sigma(\mathbf{X}))$ of $L(\sigma(\mathbf{X}))$ can be derived by computing the gradient of implicit function $\sigma(\mathbf{X})$. The approximation details are given as follows:

$$\mathbf{G}_L(\sigma(\mathbf{X})) = \mathbf{J}_\sigma(\mathbf{X})^T \mathbf{I} \mathbf{J}_\sigma(\mathbf{X}), \quad (7)$$

where $\mathbf{J}_\sigma(\mathbf{X}) \in \mathbb{R}^{|K| \times ((|U|+|I|) \times f)}$ denotes the Jacobian matrix of $\sigma(\mathbf{X})$. Then, with Hessian-free optimization, there is no need to store the large-scale Jacobian matrix $\mathbf{J}_\sigma(\mathbf{X})$ and Gauss-Newton matrix $\mathbf{G}_L(\sigma(\mathbf{X}))$. Instead, we only need to compute the Jacobian-vector product twice in each conjugate gradient iteration. The computational cost is linearly related to the storage overhead and the computation and storage of the gradient.

*C. Hessian-vector Product*

In each conjugate gradient iteration, the Hessian-vector product can be derived as follows:

$$\begin{aligned}\boldsymbol{\omega}_L(\sigma(\mathbf{X})) &= \mathbf{G}_L(\sigma(\mathbf{X}))\boldsymbol{v} \\ &= \mathbf{J}_\sigma(\mathbf{X})^\mathrm{T} \mathbf{J}_\sigma(\mathbf{X})\boldsymbol{v},\end{aligned} \tag{8}$$

where $\boldsymbol{\omega}_L(\sigma(\mathbf{X})) \in \mathbb{R}^{(|U|+|I|) \times f}$ denotes Hessian-vector product, and $\boldsymbol{v} \in \mathbb{R}^{(|U|+|I|) \times f}$ denotes the conjugate direction in each conjugate gradient iteration. Note that the initial $\boldsymbol{v}$ is a negative gradient vector, i.e., $\boldsymbol{v}^0 = -\boldsymbol{g}_E(\mathbf{X})$. The details of the Jacobian-vector product $\mathbf{J}_\sigma(\mathbf{X})\boldsymbol{v}$ calculating rules are given as:

$$\begin{aligned}\mathbf{J}_\sigma(\mathbf{X})\boldsymbol{v} &= R\left\{\sigma(\mathbf{X})_{u,i}\Big|_{(u,i)\in K}\right\} \\ &= \left(\sum_{d=1}^f (v_{u,d}x_{i,d} + x_{u,d}v_{i,d})\Big|_{(u,i)\in K}\right),\end{aligned} \tag{9}$$

where $R\{\cdot\}$ denotes the *R*-operator [33-40, 73]. Then, the Jacobian matrix $\mathbf{J}_\sigma(\mathbf{X})$ can be derived as follows:

$$\mathbf{J}_\sigma(\mathbf{X}) = \left(\frac{\partial}{\partial \mathbf{X}}\sigma(\mathbf{X})_{u,i}\Big|_{(u,i)\in K}\right), \tag{10}$$

Combining Eq. (9) and Eq. (10), the $L(\sigma(\mathbf{X}))$'s element form Hessian-vector product can be derived as follows:

$$\begin{cases}\forall u \in \boldsymbol{U}, d = 1 \sim f: \\ \boldsymbol{\omega}_L(\sigma(\mathbf{X}))_{u,d} = \sum_{i \in K_u}\left(x_{i,d}\left(\sum_{d=1}^f (v_{u,d}x_{i,d} + x_{u,d}v_{i,d})\right)\right) \\ \forall i \in \boldsymbol{I}, d = 1 \sim f: \\ \boldsymbol{\omega}_L(\sigma(\mathbf{X}))_{i,d} = \sum_{u \in K_i}\left(x_{u,d}\left(\sum_{d=1}^f (v_{u,d}x_{i,d} + x_{u,d}v_{i,d})\right)\right)\end{cases} \tag{11}$$

where $\boldsymbol{K}_u$ and $\boldsymbol{K}_i$ denote the subset w.r.t $u$ and $i$, respectively.

*D. Tikhonov Regularization Term Incorporation*

Following the computation rules of the *R*-operator, the Tikhonov regularization term of the objective function's Hessian-vector product vector $\boldsymbol{\omega}_T(\mathbf{X}) \in \mathbb{R}^{(|U|+|I|) \times f}$ in each conjugate gradient step can be calculated as:

$$\begin{aligned}\boldsymbol{\omega}_T(\mathbf{X}) &= \mathbf{H}_E(\mathbf{X})\boldsymbol{v} - \mathbf{H}_L(\sigma(\mathbf{X}))\boldsymbol{v} \\ &\Rightarrow \begin{cases}\forall u \in \boldsymbol{U}, d = 1 \sim f: \\ \boldsymbol{\omega}_T(\mathbf{X})_{u,d} = \lambda v_{u,d}|\boldsymbol{K}_u|, \\ \forall i \in \boldsymbol{I}, d = 1 \sim f: \\ \boldsymbol{\omega}_T(\mathbf{X})_{i,d} = \lambda v_{i,d}|\boldsymbol{K}_i|,\end{cases}\end{aligned} \tag{12}$$

where $|\boldsymbol{K}_u|$ denotes the volume of set $\boldsymbol{K}_u$, the same as $|\boldsymbol{K}_i|$. Then, the element form of the Hessian-vector product of Eq. (1) is as follows:

$$\begin{aligned}\boldsymbol{\omega}_E(\mathbf{X}) &\approx \mathbf{G}_E(\mathbf{X})\boldsymbol{v} \\ &\approx \boldsymbol{\omega}_L(\sigma(\mathbf{X})) + \boldsymbol{\omega}_T(\mathbf{X}),\end{aligned} \tag{13}$$

where $\boldsymbol{\omega}_E(\sigma(\mathbf{X})) \in \mathbb{R}^{(|U|+|I|) \times f}$ denotes the Hessian-vector product of Eq. (3). Its single element dependent form can be expanded as:

$$\boldsymbol{\omega}_E(\mathbf{X}) \Rightarrow \begin{cases}\forall u \in \boldsymbol{U}, d = 1 \sim f: \\ \boldsymbol{\omega}_E(\mathbf{X})_{u,d} = \sum_{i \in K_u}\left(x_{i,d}\left(\sum_{d=1}^f (v_{u,d}x_{i,d} + x_{u,d}v_{i,d})\right)\right) \\ \qquad\qquad + \lambda v_{u,d}|\boldsymbol{K}_u|, \\ \forall i \in \boldsymbol{I}, d = 1 \sim f: \\ \boldsymbol{\omega}_E(\mathbf{X})_{i,d} = \sum_{u \in K_i}\left(x_{u,d}\left(\sum_{d=1}^f (v_{u,d}x_{i,d} + x_{u,d}v_{i,d})\right)\right) \\ \qquad\qquad + \lambda v_{i,d}|\boldsymbol{K}_i|,\end{cases} \tag{14}$$

## E. Damping Term Incorporation

Considering the bilinear and non-convex nature of the objective function $E(\mathbf{X})$, its curvature approximation cannot be fully trusted. Adding a large enough damping term in the curvature matrix can avoid ill-conditioned issues and improve convergence performance [33-39, 73]. The single-factor form's Hessian-vector product vector $\omega_E(\sigma(\mathbf{X}))$ incorporating the damping term can be expanded as follows:

$$\omega_E(\mathbf{X}) \approx (\mathbf{G}_E(\mathbf{X}) + \gamma \mathbf{I})\mathbf{v}$$
$$\Rightarrow \begin{cases} \forall u \in \mathbf{U}, d = 1 \sim f: \\ \omega_E(\mathbf{X})_{u,d} = \sum_{i \in K_u}\left(x_{i,d}\left(\sum_{d=1}^{f}(v_{u,d}x_{i,d} + x_{u,d}v_{i,d})\right)\right) \\ \qquad\qquad + \lambda v_{u,d}|K_u| + \gamma v_{u,d}, \\ \forall i \in \mathbf{I}, d = 1 \sim f: \\ \omega_E(\mathbf{X})_{i,d} = \sum_{u \in K_i}\left(x_{u,d}\left(\sum_{d=1}^{f}(v_{u,d}x_{i,d} + x_{u,d}v_{i,d})\right)\right) \\ \qquad\qquad + \lambda v_{i,d}|K_i| + \gamma v_{i,d}, \end{cases} \quad (15)$$

where $\gamma$, the damping term coefficient, is regarded as the regularization term of the curvature matrix $\mathbf{G}_E(\mathbf{X})$, balancing the first-order approximation and second-order approximation.

## F. Gradient Estimation via PID Controller Principle

Before initiating the conjugate gradient method, it is necessary to compute the negative gradient $-\mathbf{g}_E(\mathbf{X})$ as the initial conjugate direction vector $\mathbf{v}$. Integrating the principles of a PID controller, represented by Eq. (5), we derive the estimated negative gradient $-\tilde{\mathbf{g}}_E(\mathbf{X})$ for Eq. (3) as follows:

$$\begin{cases} \forall u \in \mathbf{U}, d = 1 \sim f: \\ -\tilde{\mathbf{g}}_E(\mathbf{X})_{u,d}^t = \sum_{i \in K_u}\left(\tilde{e}_{u,i}^t x_{i,d} - \lambda x_{u,d}\right), \\ \forall i \in \mathbf{I}, d = 1 \sim f: \\ -\tilde{\mathbf{g}}_E(\mathbf{X})_{i,d}^t = \sum_{u \in K_i}\left(\tilde{e}_{u,i}^t x_{u,d} - \lambda x_{i,d}\right), \end{cases} \quad (16)$$

## G. Update Rule

Based on the above analysis, the increment vector can be derived via multiple conjugate gradient computations. At $t$-th epoch, the proposed method's update rule is as follows:

$$\begin{cases} \Delta\mathbf{X}^t \overset{\text{Mutiple Conjugate Gradient Steps}}{\Longleftarrow} (\mathbf{G}_E(\mathbf{X})^t + \gamma\mathbf{I})\Delta\mathbf{X}^t + \tilde{\mathbf{g}}_E(\mathbf{X})^t \leq \tau\mathbf{I}, \\ \mathbf{X}^{t+1} = \mathbf{X}^t + \Delta\mathbf{X}^t, \end{cases} \quad (17)$$

where constant $\tau$ denotes the tolerance in the conjugate gradient that terminates its iterations [33-39, 73].

## IV. EXPERIMENTS

In this section, all experiment details are given.

## A. General Settings

TABLE II. INVOLVED DATASETS

| No. | Name | $|U|$ | $|I|$ | $|K|$ | Density |
|---|---|---|---|---|---|
| D1 | ML-1M | 6,040 | 3,952 | 1,000,290 | 4.19% |
| D2 | IMDB | 2,113 | 10,197 | 855,598 | 3.97% |
| D3 | Personality | 854 | 13,012 | 458,970 | 4.13% |
| D4 | LTS | 1820 | 35,196 | 1,028,751 | 1.61% |

**Datasets.** Four open-access benchmark HDI datasets are involved in this section, i.e., MovieLens 1M (ML-1M) [99], IMDB [101], Personality [103], and Learning to Sets (LTS) [105]. The details of all benchmark datasets are summarized in Table II.

**Evaluation metric.** We adopt the root mean square error (RMSE) to evaluate the prediction performance of the proposed method and other competitors. The lower the RMSE value, the better prediction performance for missing elements in the HDI matrix [59-98, 100, 102, 104, 106, 108, 109].

$$RMSE = \sqrt{\frac{\sum_{r_{u,i} \in \Omega}\left(r_{u,i} - \sum_{d=1}^{f} x_{u,d}x_{i,d}\right)^2}{|\Omega|}} \quad (18)$$

where $\Omega$ denotes the evaluation set.

**Competitors.** The proposed PSLF model is compared to the following four state-of-the-art advanced-optimizer-based LFA models and summarized in Table III.

1) **An SGD-based LFA model (M1)** [24-28]**:** The vanilla SGD is a default optimizer for any representation learning model including the vanilla LFA model.

2) **An Adam-based LFA model (M2)** [29]**:** Adam is an extension of SGD optimizer that incorporates tiny curvature information (aka second-order momentum) to rescale first-order momentum. This optimizer well performs in optimizing non-convex functions.

3) **An SAM-based LFA model (M3)** [107]**:** Sharpness-Aware Minimization (SAM), an extension SGD optimizer, seeks decision parameters optimized in a flatter minima to enhance the model's generalization.

4) **An SLF model (M4)** [33]: A second-order LFA model via the Hessian-vector product method.

5) **A PSLF model (M5):** The proposed PSLF model in this paper.

TABLE III. DETAILS OF COMPARED MODELS

| No. | Name | Type |
| --- | --- | --- |
| M1 | SGD-based LFA | First-order |
| M2 | Adam-based LFA | Adaptive-gradeint |
| M3 | SAM-based LFA | First-order |
| M4 | SLF | Second-order |
| M5 | PSLF | Second-order |

**Setting Strategy.** The detailed information regarding all benchmark datasets and the settings of the compared models is as follows:

1) *Environment Configuration:* All experiments are finished on a Linux server with Ubuntu 20.04.6 LTS 64-bit, two-way Intel Xeon Silver 4214R 2.4 GHz, 12 cores and 24 threads for each way CPU, and 128 GB RAM. Moreover, all tested models are running on OpenJDK 11 LTS.

2) *Dataset Partition Strategy:* All benchmark datasets are randomly divided into three subsets: training set, validation set, and test set, with a partition ratio of 60% for training, 20% for validation, and 20% for testing.

3) *Hyper-parameters Optimization:* According to [19, 20] the initial latent factor matrices are sampled from $U(0, 0.04)$. The dimension of latent space $f$ is set to 20. The optimal hyperparameter combinations for all tested models are determined through a grid search conducted on the validation set. Based on [6, 8, 9, 12, 13, 15, 36], the search range of learning rate for M1-M3 are set as $[2^{-8}, 2^{-9},…, 2^{-12}]$, $[10^{-1}, 10^{-2},…, 10^{-5}]$, $[2^{-8}, 2^{-9},…,2^{-12}]$, respectively. The learning rate for M4 and M5 are set to 1. The regularization term coefficient $\lambda$ for all tested models is set within the same search range, i.e., $[0.01, 0.02,…, 0.09]$. For M2, the damping term $\gamma$ is set to $10^{-8}$, the first-order momentum and second-order momentum coefficients are set to recommended values, i.e., 0.9 and 0.999, respectively. For M3, the search range of hyper-parameter $\rho$ is set as $[2^{-1}, 2^{-2},…, 2^{-10}]$. M4 and M5's conjugate gradient terminated conditions, i.e., tolerance $\tau$, is set to 100. The search range of damping term coefficient $\gamma$ is set as $[10, 30, 60,…, 300]$. For M5, the hyper-parameters of the PID controller, i.e., $K_P$, $K_I$, and $K_D$, are set to 1.5, 0.005, and 0.05, respectively.

4) *Terminate Condition:* All tested models share the same termination strategy. The maximum training epoch is 500. The early stop epoch is set to 10. In other words, if the validation set's RMSE value worsens compared to the optimal validation error over 10 epochs, the training process will be stopped, and the model's generalization error will be evaluated on the test set using the optimal latent factor matrices.

*B. Comparison Results*

The comparison results on four open-access benchmark datasets are summarized in Table IV. From Table IV, we can make the following observations:

**(1) Refining learning error via incorporating the principle of a PID controller can improve prediction accuracy for the HDI matrix's missing data.** As shown in Table IV, the average RMSE values for M5 are 0.85082, 0.77026, 0.86237, and 0.81855 on D1-D4, respectively. For M4, the average RMSE values are 0.85132, 0.77096, 0.86382, and 0.81962, respectively. It is noteworthy that a lower RMSE value indicates better generalization ability for the LFA model in this paper. The PID-incorporated SLF model M5 exhibits lower RMSE values on D1-D4 compared to the vanilla SLF model. Specifically, on D1-D4, M5's average RMSE values are lower than M1-M4 0.06%, 0.09%, 0.17%, and 0.13%, respectively.

**(2) Incorporating the principle of a PID controller can accelerate the convergence rate.** For instance, from Table IV, M5's average time consumption and training epochs are 29 seconds and 36 epochs on D1, respectively, which are about 9.38% and 29.41% lower than M4's 32 seconds and 51 epochs, respectively. On D3, M5's time cost and iterations are 101 seconds and 43 epochs, respectively, which are about 28.87% and 61.95% lower than M4's 142 seconds and 113 iterations, respectively. Similar results can be observed across the other datasets as well, as shown in Table IV.

**(3) PSLF demonstrates superior performance over all tested first-order-based LFA models in low-rank representation ability.** By incorporating curvature information, the PSLF model achieves the lowest RMSE value on D1-D4 compared to the first-order-based LFA models. For instance, on the D1 dataset, M5's RMSE values are lower than those of M1 by 0.13%, M2 by 0.11%, and M3 by 0.13%, respectively. Similarly, on D3, M5's RMSE values are lower than those of M1 by 0.23%, M2 by 0.17%, and M3 by 0.17%, respectively. These trends are consistent across the other datasets, as shown in Table IV.

**(4) M5 achieves better computational efficiency compared to first-order-based LFA competitors.** According to Table IV, the proposed M5, a second-order-based LFA model, converges to local minima faster than the tested gradient-based LFA models. For example, based on the results from D1 in Table IV, M5 requires only 29 seconds to converge, whereas the first-order-based LFA models require more time to converge. Specifically, M1 needs 132 seconds, M2 needs 1059 seconds, and M3 needs 172 seconds. Additionally, the convergence epoch for M5 is 47 rounds, whereas the first-order-based LFA models require more iterations to converge to a stable point. In detail, M1 requires 469 rounds, M2 needs 436 rounds, and M3 needs 389 rounds to converge to a stable point, respectively. Similar observations can be made for D2-D4 from Table IV.

TABLE IV. PERFORMANCE BENCHMARK ON D1-D5

| Datasets | Model | RMSE | Time (Sec) | Epoch |
|---|---|---|---|---|
| D1 | M1 | 0.85193±0.00082 | 132±4 | 469±4 |
| | M2 | 0.85173±0.00058 | 1059±9 | 436±4 |
| | M3 | 0.85190±0.00080 | 173±14 | 489±11 |
| | M4 | 0.85132±0.00101 | 32±0 | 51±3 |
| | M5 | **0.85082±0.00106** | **29±2** | **36±3** |
| D2 | M1 | 0.77141±0.00082 | 107±2 | 439±10 |
| | M2 | 0.77137±0.00104 | 965±18 | 458±1 |
| | M3 | 0.77137±0.00076 | 137±9 | 453±7 |
| | M4 | 0.77096±0.00083 | 35±1 | 82±1 |
| | M5 | **0.77026±0.00067** | **31±0** | **39±3** |
| D3 | M1 | 0.86433±0.00081 | 107±3 | 349±17 |
| | M2 | 0.86384±0.00062 | 1163±63 | 468±3 |
| | M3 | 0.86388±0.00031 | 144±12 | 351±27 |
| | M4 | 0.86382±0.00105 | 142±2 | 113±0 |
| | M5 | **0.86237±0.00129** | **101±6** | **43±1** |
| D5 | M1 | 0.82045±0.00009 | 37±4 | 339±3 |
| | M2 | 0.82120±0.00008 | 535±3 | 489±4 |
| | M3 | 0.82032±0.00017 | 58±3 | 340±13 |
| | M4 | 0.81962±0.00014 | 45±4 | 95±1 |
| | M5 | **0.81855±0.00008** | **27±2** | **28±2** |

*C. Summary*

In this section, we compared the proposed PSLF model with advanced optimizer-based LFA models in terms of generalization and convergence rate. Empirical results demonstrate that the PSLF model outperforms its competitors in predicting missing data within the HDI matrix and achieving faster convergence rates.

## V. CONCLUSION

In this paper, we proposed a PID controller-incorporated SLF model to improve the performance of representing HDI interaction data. The incorporation of the PID controller principles significantly enhances prediction accuracy by refining learning error estimation, as evidenced by lower RMSE values across multiple datasets. Additionally, this approach accelerates the convergence rate, resulting in reduced training time and fewer iterations compared to the vanilla second-order-based latent factor model and state-of-the-art optimizer-based latent factor models. Experimental results demonstrate that the proposed model not only achieves superior generalization performance but also offers better computational efficiency. Overall, the integration of a PID controller with second-order-based latent factor provides a robust and efficient solution for representing HDI data, outperforming existing state-of-the-art optimizer-based latent factor models.